\documentclass{article}
\usepackage{ijcai19}

\usepackage{times}
\usepackage{soul}
\usepackage{url}
\usepackage[hidelinks]{hyperref}
\usepackage[utf8]{inputenc}
\usepackage[small]{caption}
\usepackage{graphicx}
\usepackage{amsmath}
\usepackage{booktabs}
\usepackage{mathtools}
\usepackage{multirow}
\usepackage[]{algorithm}
\usepackage{algorithmic}
\usepackage{comment}
\DeclareMathOperator*{\argmax}{arg\,max}
\DeclareMathOperator*{\argmin}{arg\,min}

\newlength\myindent
\title{Adversarial Exploitation of Policy Imitation}

\author{
Vahid Behzadan\footnote{Contact Author}\And
William H. Hsu
\affiliations
Kansas State University\\
\emails
\{behzadan, bhsu\}@ksu.edu
}

\begin{document}

\maketitle

\begin{abstract}
This paper investigates a class of attacks targeting the confidentiality aspect of security in Deep Reinforcement Learning (DRL) policies. Recent research have established the vulnerability of supervised machine learning models (e.g., classifiers) to model extraction attacks. Such attacks leverage the loosely-restricted ability of the attacker to iteratively query the model for labels, thereby allowing for the forging of a labeled dataset which can be used to train a replica of the original model. In this work, we demonstrate the feasibility of exploiting imitation learning techniques in launching model extraction attacks on DRL agents. Furthermore, we develop proof-of-concept attacks that leverage such techniques for black-box attacks against the integrity of DRL policies. We also present a discussion on potential solution concepts for mitigation techniques.
\end{abstract}

\section{Introduction}
Recent research have established the vulnerability of supervised machine learning models (e.g., classifiers) to model extraction attacks\cite{tramer2016stealing}. Such attacks leverage the loosely-restricted ability of the attacker to iteratively query the model for labels, thereby allowing for the forging of a labeled dataset which can be used to train a replica of the original model. Model extraction is not only a serious risk to the protection of intellectual property, but also a critical threat to the integrity of the model. Recent literature\cite{behzadan2017vulnerability} report that the replicated model may facilitate the discovery and crafting of adversarial examples which are transferable to the original model.

Inspired by this area of research, this work investigates the feasibility and impact of model extraction attacks on DRL agents. The adversarial problem of model extraction can be formally stated as the replication of a DRL policy based on observations of its behavior (i.e., actions) in response to changes in the environment (i.e., state). This problem closely resembles that of imitation learning\cite{hussein2017imitation}, which refers to the acquisition of skills or behaviors by observing demonstrations of an expert performing those skills. Typically, the settings of imitation learning are concerned with learning from human demonstrations. However, it is straightforward to deduce that the techniques developed for those settings may also be applied to learning from artificial experts, such as DRL agents. Of particular relevance to this research is the emerging area of Reinforcement Learning with Expert Demonstrations (RLED)\cite{piot2014boosted}. The techniques of RLED aim to minimize the effect of modeling imperfections on the efficacy of the final RL policy, while minimizing the cost of training by leveraging the information available demonstrations to reduce the search space of the policy. 

Accordingly, we hypothesize that the techniques developed for RLED may be maliciously exploited to replicate and manipulate DRL policies. To establish the validity of this hypothesis, we investigate the feasibility of RLED techniques in utilizing limited passive (i.e., non-interfering) observations of a DRL agent to replicate its policy with sufficient accuracy to facilitate attacks on their integrity. To develop proof-of-concept attacks, we study the adversarial utility of adopting a recently proposed RLED technique, known as Deep Q-Learning from Demonstrations (DQfD)\cite{hester2018deep} for black-box state-space manipulation attacks, and develop two attack mechanisms based on this technique. Furthermore, we present a discussion on potential mitigation techniques, and present a solution concept for defending against policy imitation attacks.

The remainder of this paper is organized as follows: Sectio \ref{sec:DQfD} presents an overview of the DQfD algorithm used in this study for adversarial imitation. Section \ref{sec:ExtractionPoC1} proposes the first proof-of-concept black-box attack based on imitated policies, and presents experimental evaluation of its feasibility and performance. Section \ref{sec:transfer} studies the transferability of adversarial examples between replicated and the original policies as a second proof-of-concept attack technique. The paper concludes with a discussion on potential mitigation techniques and a solution concept in Section \ref{sec:discussion}.

\section{Deep Q-Learning from Demonstrations (DQfD)}
\label{sec:DQfD}
The DQfD technique\cite{hester2018deep} aims to overcome the inaccuracies of simulation environments and models of complex phenomenon by enabling DRL agents to learn as much as possible from expert demonstrations before training on the real system. More formally, the objective of this ``pre-training'' phase is to learn an imitation of the expert's behavior with a value function that is compatible with the Bellman equation, thereby enabling the agent to update this value function via TD updates through direct interaction with the environment after the pre-training stage. To achieve such an imitation from limited demonstration data during pre-training, the agent trains on sampled mini-batches of demonstrations to train a deep neural network model in a supervised manner. However, the training objective of this model in DQfD is the minimization of a hybrid loss, comprised of the following components:
\begin{enumerate}
    \item 1-step double Q-learning loss $J_{DQ}(Q)$,
    \item Supervised large margin classification loss $J_E(Q) = \max_{a\in A} [Q(s,a) + l(a_E,a)] - Q(s,a_E)$, where $a_E$ is the expert's action in state $s$ and $l(a_E,a)$ is a margin function that is positive if $a\neq a_E$, and is 0 when $a=a_E$.
    \item ($n=10$)-step Return: $r_t+\gamma r_{t+1}+...+\gamma^{n-1}r_{t+n-1}+\max_a \gamma^n Q(s_{t+n},a)$.
    \item L2 regularization loss: $J_{L2}(Q)$ 
\end{enumerate}
The total loss is given by:
\begin{equation}
    J(Q) = J_{DQ}(Q) + \lambda_1 J_n(Q) + \lambda_2 J_E(Q) + \lambda_3 J_{L2}(Q)
\end{equation}
where $\lambda$ factors provide the weighting between the losses.

After the pre-training phase, the agent begins interacting with the system and collecting self-generated data, which is added to the replay buffer $D^{replay}$. Once the buffer is full, the agent only overwrites the self-generated data and leaves the demonstration data untouched for use in the coming updates of the model. The complete training procedure for DQfD is presented in Algorithm \ref{alg:ValueApproximation}.

\begin{algorithm}[h] 
\caption{Deep Q-learning from Demonstrations (DQfD)}
\label{alg:ValueApproximation} 
\begin{algorithmic} 
    \STATE Inputs: $D^{replay}$ initialized with demonstration data, randomly initialized weights for the behavior network $\theta$, randomly initialized weights for the target network $\theta'$, updating frequency of the target network $\tau$, number of pre-training gradient updates $k$
    \FOR{steps $t\in \{1,2,...,k\}$}
        \STATE Sample a mini-batch of $n$ transitions from $D^{replay}$ with prioritization
        \STATE Calculate loss $J(Q)$ based on target network
        \STATE Perform a gradient descent step to update $\theta$
        \IF{$t$ mod $\tau = 0$}
            \STATE $\theta'\leftarrow \theta$
        \ENDIF
    \ENDFOR
    \FOR{steps $t\in \{1,2,...\}$}
        \STATE Sample action from behavior policy $a ~ \pi^{\epsilon Q_\theta}$
        \STATE Apply action $a$ and observe $(s', r)$
        \STATE Store $(s,a,r,s')$ into $D^{replay}$, overwriting oldest self-generated transition if over capacity
        \STATE Sample a mini-batch of $n$ transitions from $D^{replay}$ with prioritization
        \STATE Calculate loss $J(Q)$ using target network
        \STATE Perform a gradient descent step to update $\theta$
        \IF{$t$ mod $\tau = 0$}
            \STATE $\theta'\leftarrow \theta$
        \ENDIF
        \STATE $s\leftarrow s'$
    \ENDFOR
\end{algorithmic}
\end{algorithm}

\section{Adversarial Policy Imitation for Black-Box Attacks}
\label{sec:ExtractionPoC1}
Consider an adversary who aims to maximally reduce the cumulative discounted return ($R(T)$) of a target DRL agent by manipulating the behavior of the target's policy $\pi(s)$ via perturbing its observations. The adversary is also constrained to minimizing the total cost of perturbations given by $C_{adv}(T) = \sum_{t=t0}^{T} c_{adv}(t)$, where $c_{adv}(t) = 1$ if the adversary perturbs the state at time $t$, and $c_{adv}(t) = 0$ otherwise. 

The adversary is unaware of $\pi(s)$ and its parameters. However, it has access to a replica of the target's environment (e.g., the simulation environment). Also, for any state transition $(s, a)\rightarrow s')$, the adversary can perfectly observe the target's reward signal $r_(s, a, s')$, and is able to observe the behavior of $\pi(s)$ in response to each state $s$. Furthermore, the adversary is able to manipulate its target's state observations, but not its reward signal. Also, it is assumed that all targeted perturbations of the adversary are successful. 

To study the feasibility of imitation learning as an approach to this adversarial problem, we consider the first step of the adversary to be the imitation of $\pi(s)$ via DQfD to learn an imitated policy $\Tilde{\pi}$. With this imitation at hand, the attack problem can be reformulated to finding an optimal adversarial control policy $\pi_{adv}(s)$, where the control actions are two-fold: whether to perturb the current state to induce the worst possible action (i.e., $\argmin_a Q(s, a)$ or to leave the state unperturbed. This setting allows for the direct adoption of the DRL-based technique proposed in [awaiting appearance on Arxiv\footnote{http://www.vbehzadan.com/drafts/RobustBenchmark.pdf}] for resilience benchmarking of DRL policies. In this technique, the test-time resilience of a policy $\pi^*$ to state-space perturbations is obtained via an adversarial DRL training procedure, outlined as follows:
\begin{enumerate}
    \item Train the adversarial agent against the target following $\pi$ in its training environment according to the reward assignment process outlined in Algorithm \ref{alg:resilience}. report the optimal adversarial return $R_{perturbed}^*$ and the maximum adversarial regret $R^*_{adv}(T)$, which is the difference between maximum achievable return by the target $\pi$ and its minimum achieved return from actions of adversarial policy. 
    \item Apply the adversarial policy against the target in $N$ episodes, record total cost $C_{adv}$ for each episode,
    \item Report the average of $C_{adv}$ over $N$ episodes as the mean test-time resilience of $\pi$ in the given environment.
\end{enumerate}. While the original technique is dependent on the availability of target's optimal state-action value function, we propose to replace this function with the $Q$-function obtained from DQfD imitation of the target policy, denoted by $\Tilde{Q}$. 

\ref{alg:resilience}:
\begin{algorithm} 
\caption{Reward Assignment in Adversarial DRL for Measuring Adversarial Resilience} 
\label{alg:resilience} 
\begin{algorithmic} 
    \REQUIRE Target policy $\pi^*$, Perturbation cost function $c_{adv}(., .)$, Maximum achievable score $R_{max}$, Optimal state-action value function $Q^*(.,.)$, Current adversarial policy $\pi^{adv}$, Current state $s_t$, Current count of adversarial actions $AdvCount$, Current score $R_t$
    \STATE Set ToPerturb $\leftarrow \pi^{adv}(s_t)$
    \IF{ToPerturb is False}
        \STATE $a_t \leftarrow \pi^*(s_t)$
        \STATE ${Reward} \leftarrow 0$
    \ELSE
        \STATE $a'_t \leftarrow \argmin_a Q^*(s_t, a)$
        \STATE ${Reward} \leftarrow - c_{adv}(s_t, a'_t)$
    \ENDIF
    \IF{either $s_t$ or $s'_t$ is terminal}
        \STATE ${Reward} += (R_{max} - R_t)$
    \ENDIF
\end{algorithmic}
\end{algorithm}

With the imitated state-action value function $\Tilde{Q}$ at hand, the adversarial policy can be trained as a DRL agent with the procedure outlined in [awaiting appearance on Arxiv\footnote{http://www.vbehzadan.com/drafts/RobustBenchmark.pdf}]. The proposed attack procedure is summarized as follows:
\begin{enumerate}
    \item Observe and record $N$ interactions $(s_t, a_t, s_{t+1}, r_{t+1})$ of the target agent with the environment.
    \item Apply DQFD to learn an imitation of the target policy $\pi(s)$ and $Q^*$, denoted by $\Tilde{\pi}$ and $\Tilde{Q}$, respectively.
    \item Train adversarial policy $\pi_{adv}(s)$ with Algorithm \ref{alg:resilience}, using $\Tilde{Q}$ as an approximation of target's $Q^*$.
    \item Apply adversarial policy to the target environment.
\end{enumerate}

\subsection{Experiment Setup}
We consider a DQN-based adversarial agent, aiming to learn an optimal adversarial state-perturbation policy to minimize the return of its targets, consisting of DQN, A2C, and PPO2 policies trained in the CartPole environment. The architecture and hyperparameters of the adversary and its targets are the same as those detailed in [awaiting appearance on Arxiv\footnote{http://www.vbehzadan.com/drafts/RobustBenchmark.pdf}]. The adversary employs a DQfD agent to learn an imitation of each target, the hyperparameters of which are provided in Table \ref{Table:DQfD}.

\begin{table}[h]
\centering
\begin{tabular}{|c|c|}
\hline
Pretraining Steps          & 5000 \\ \hline
Large Margin               & 0.8  \\ \hline
Imitation Loss Coefficient & 1    \\ \hline
Target Update Freq.        & 1000 \\ \hline
n-steps                    & 10   \\ \hline
$\gamma$                   & 0.99 \\ \hline
\end{tabular}
\caption{Parameters of DQfD Agent}
\label{Table:DQfD}
\end{table}

\subsection{Results}
\begin{figure}[H]
	
	\centering
	
	\includegraphics[width = 0.8\columnwidth]{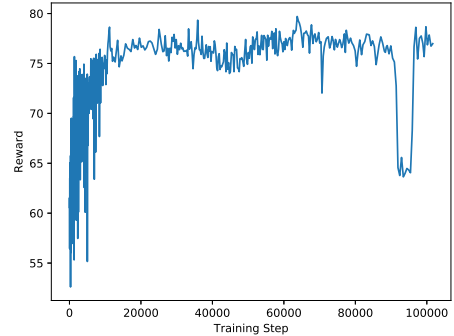}
	
	\caption{DQfD Training Progress on DQN Policy with 5k demonstrations}
	
	\label{fig:DQfDDQN5k}
	
\end{figure}

\begin{figure}[H]
	
	\centering
	
	\includegraphics[width = 0.8\columnwidth]{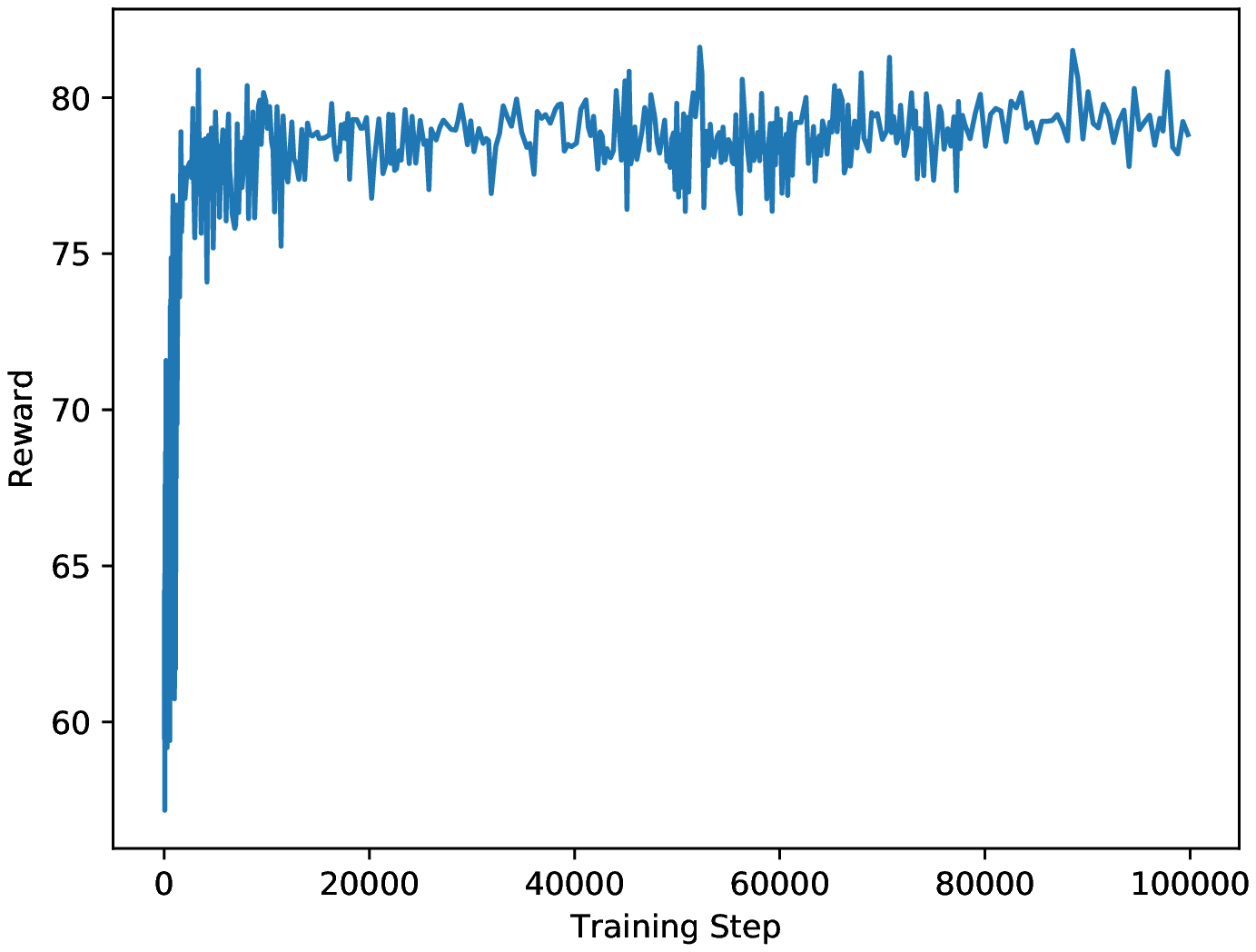}
	
	\caption{DQfD Training Progress on A2C Policy with 5k demonstrations}
	
	\label{fig:DQfDA2C5k}
	
\end{figure}

\begin{figure}[H]
	
	\centering
	
	\includegraphics[width = 0.8\columnwidth]{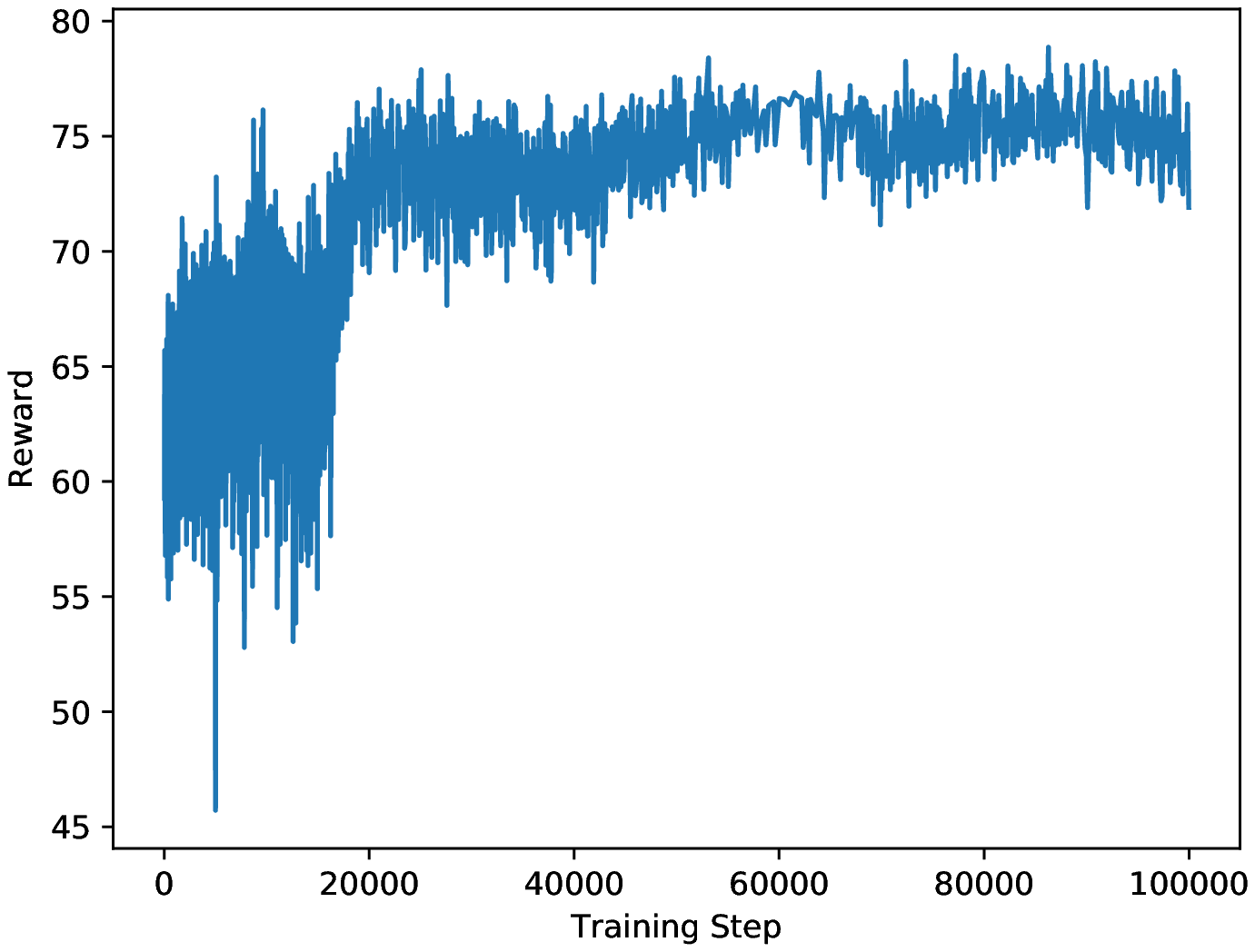}
	
	\caption{DQfD Training Progress on PPO2 Policy with 5k demonstrations}
	
	\label{fig:DQfDPPO5k}
	
\end{figure}

Figures \ref{fig:DQfDDQN5k} -- \ref{fig:DQfDPPO5k} illustrate the first 100000 training steps of DQfD from 5000 observations obtained from DQN, A2C, and PPO2 policies in CartPole. While this limited window of training is not long enough for convergence to an optimal policy in CartPole, the following results demonstrate its sufficiency for deriving adversarial perturbation policies for all three targets.

\begin{figure}[H]
	
	\centering
	
	\includegraphics[width = 0.8\columnwidth]{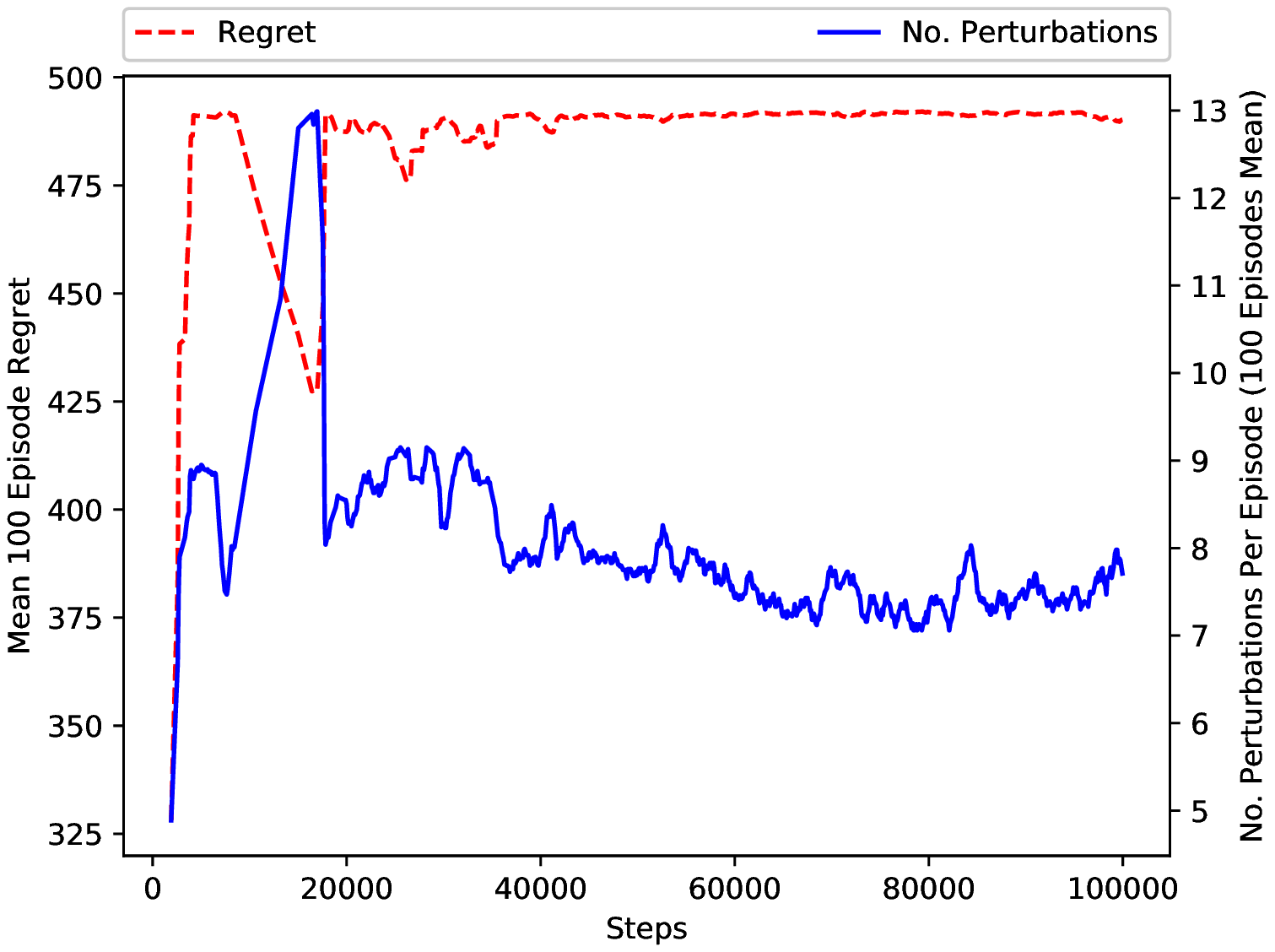}
	
	\caption{Adversarial Training Progress on DQN Policy with 5k demonstrations}
	
	\label{fig:DQN5k}
	
\end{figure}

\begin{figure}[H]
	
	\centering
	
	\includegraphics[width = 0.8\columnwidth]{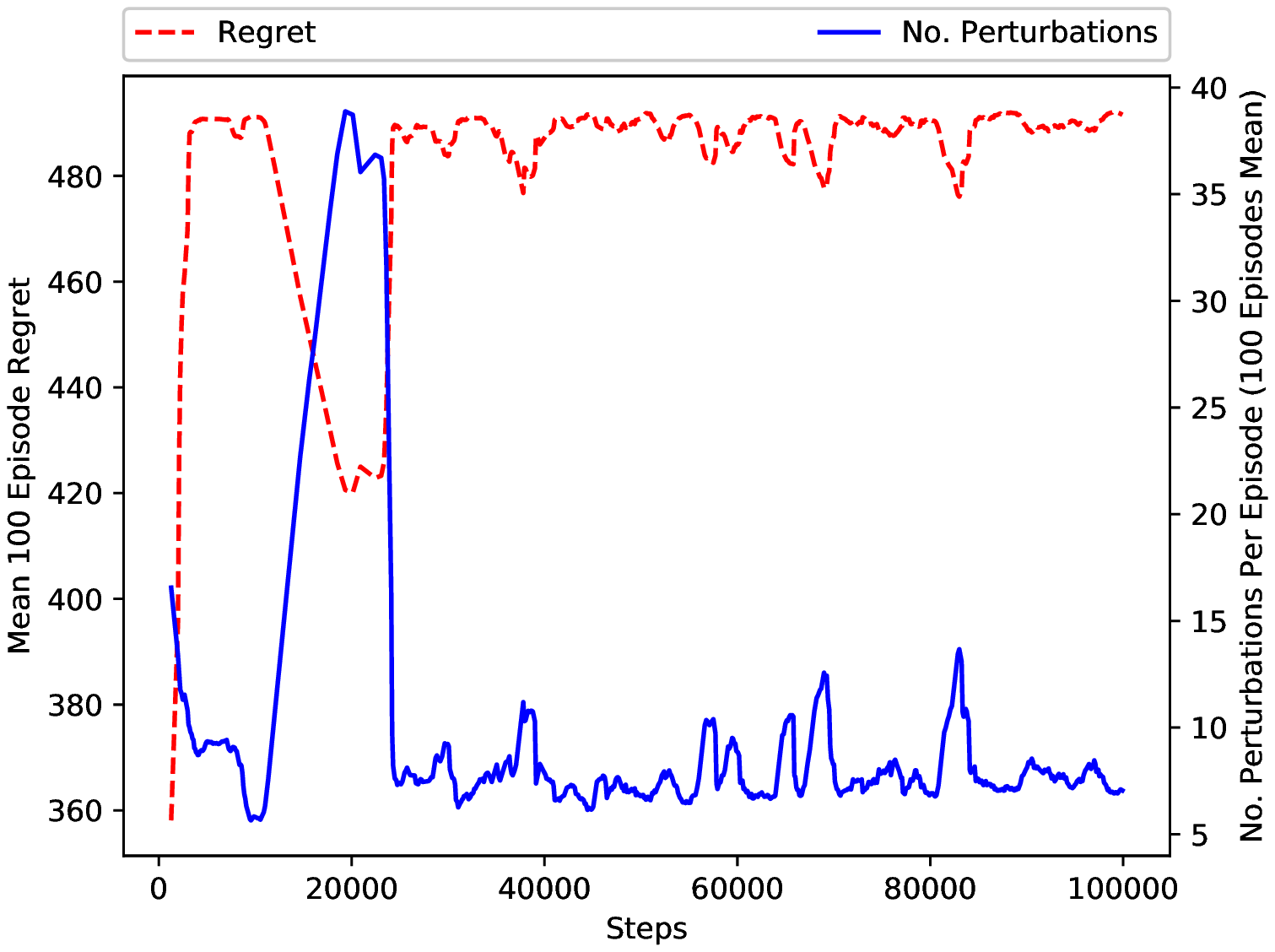}
	
	\caption{Adversarial Training Progress on DQN Policy with 2.5k demonstrations}
	
	\label{fig:DQN2k}
	
\end{figure}

\begin{figure}[H]
	
	\centering
	
	\includegraphics[width = 0.8\columnwidth]{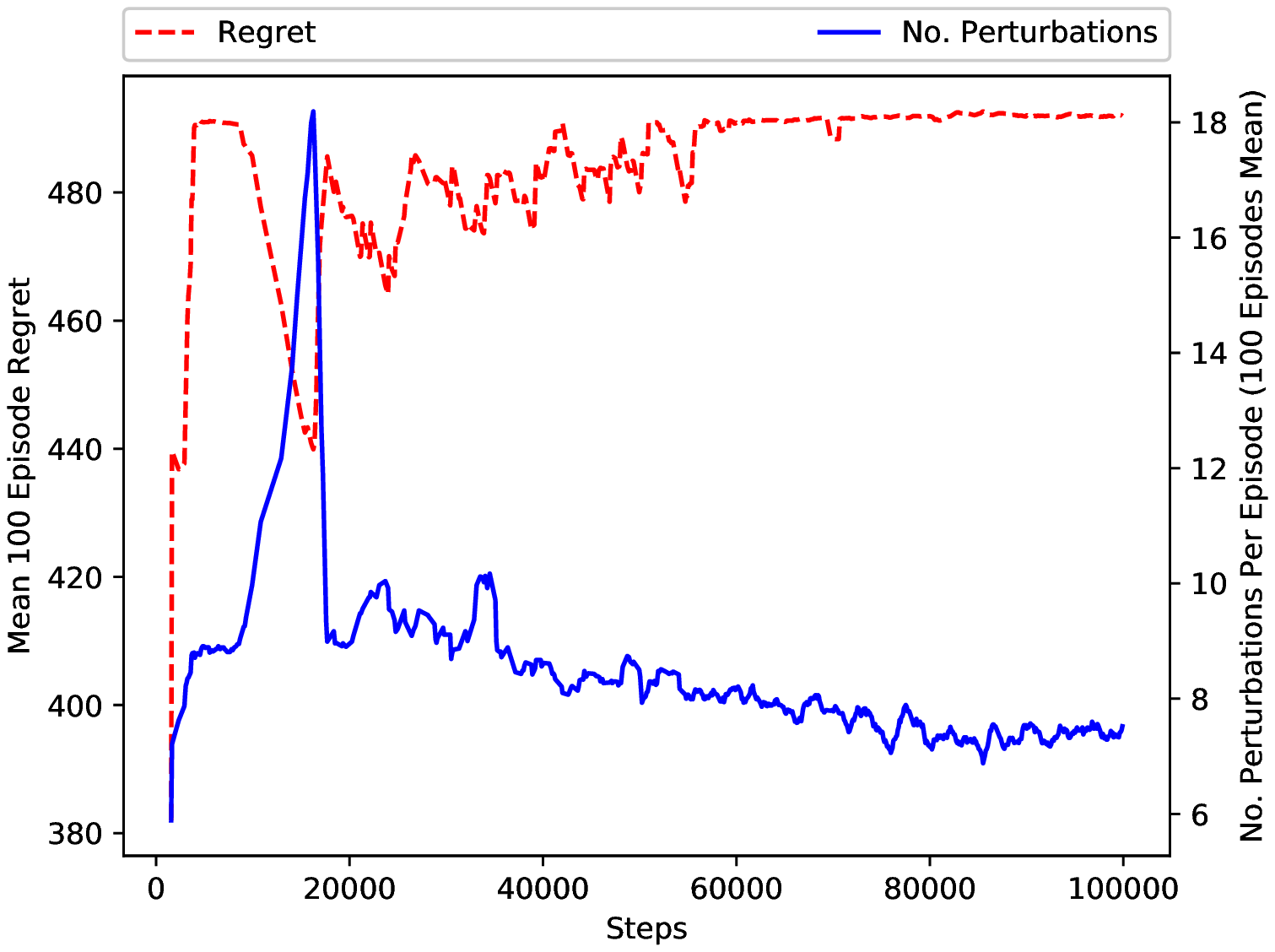}
	
	\caption{Adversarial Training Progress on DQN Policy with 1k demonstrations}
	
	\label{fig:DQN1k}
	
\end{figure}

\begin{figure}[H]
	
	\centering
	
	\includegraphics[width = 0.8\columnwidth]{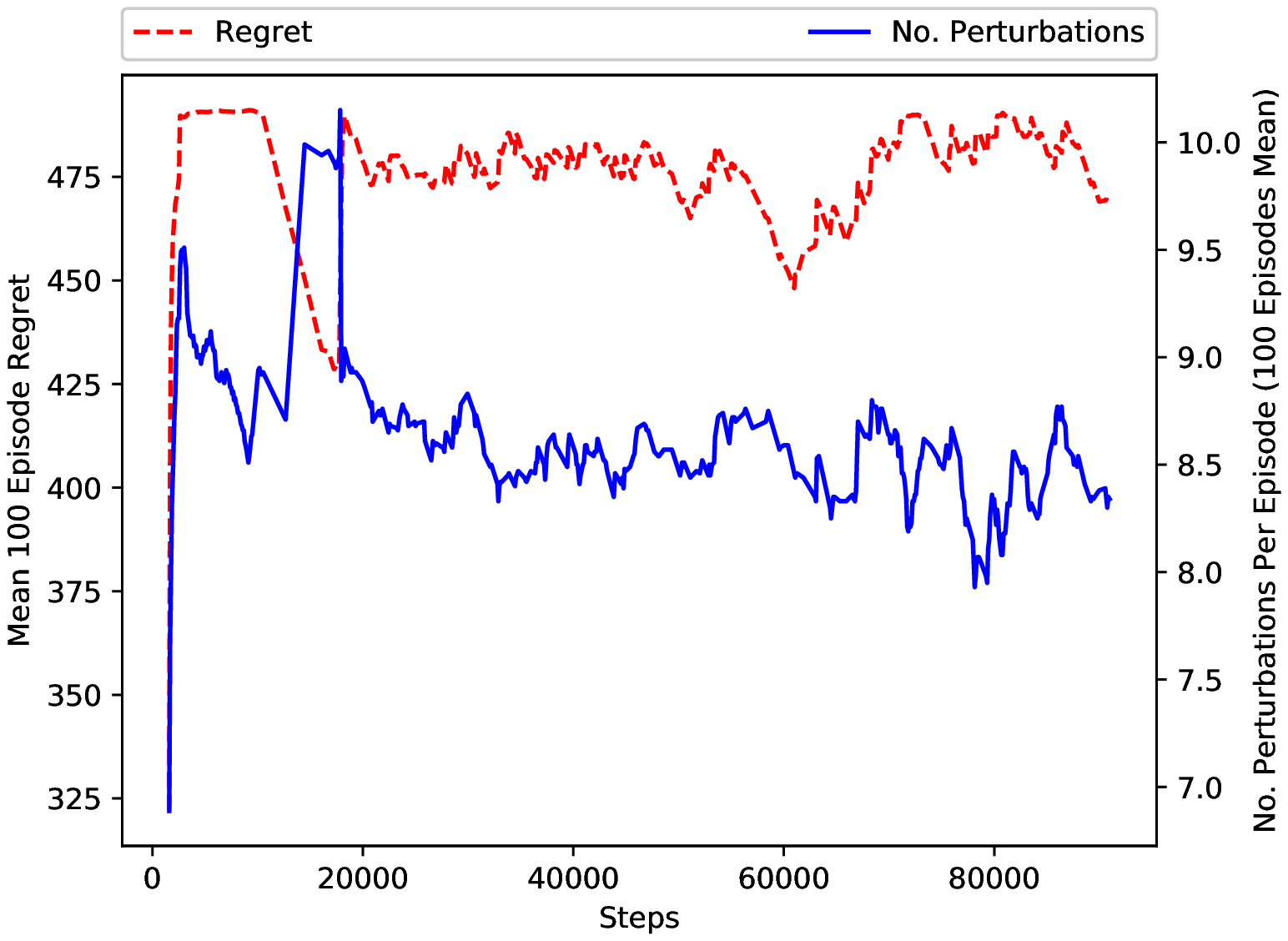}
	
	\caption{Adversarial Training Progress on A2C Policy with 5k demonstrations}
	
	\label{fig:A2C5k}
	
\end{figure}

\begin{figure}[H]
	
	\centering
	
	\includegraphics[width = 0.8\columnwidth]{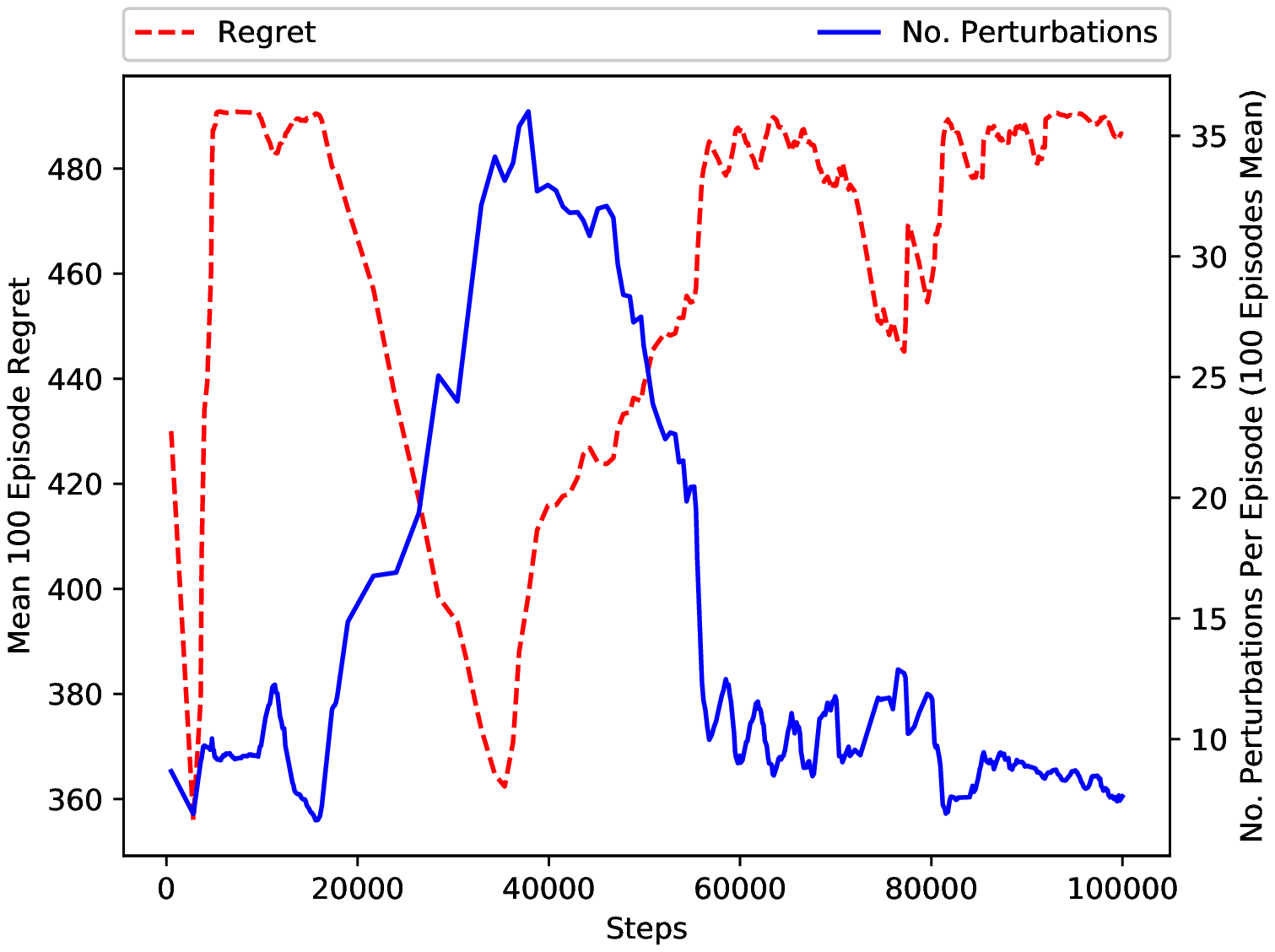}
	
	\caption{Adversarial Training Progress on A2C Policy with 2.5k demonstrations}
	
	\label{fig:A2C2k}
	
\end{figure}

\begin{figure}[H]
	
	\centering
	
	\includegraphics[width = 0.8\columnwidth]{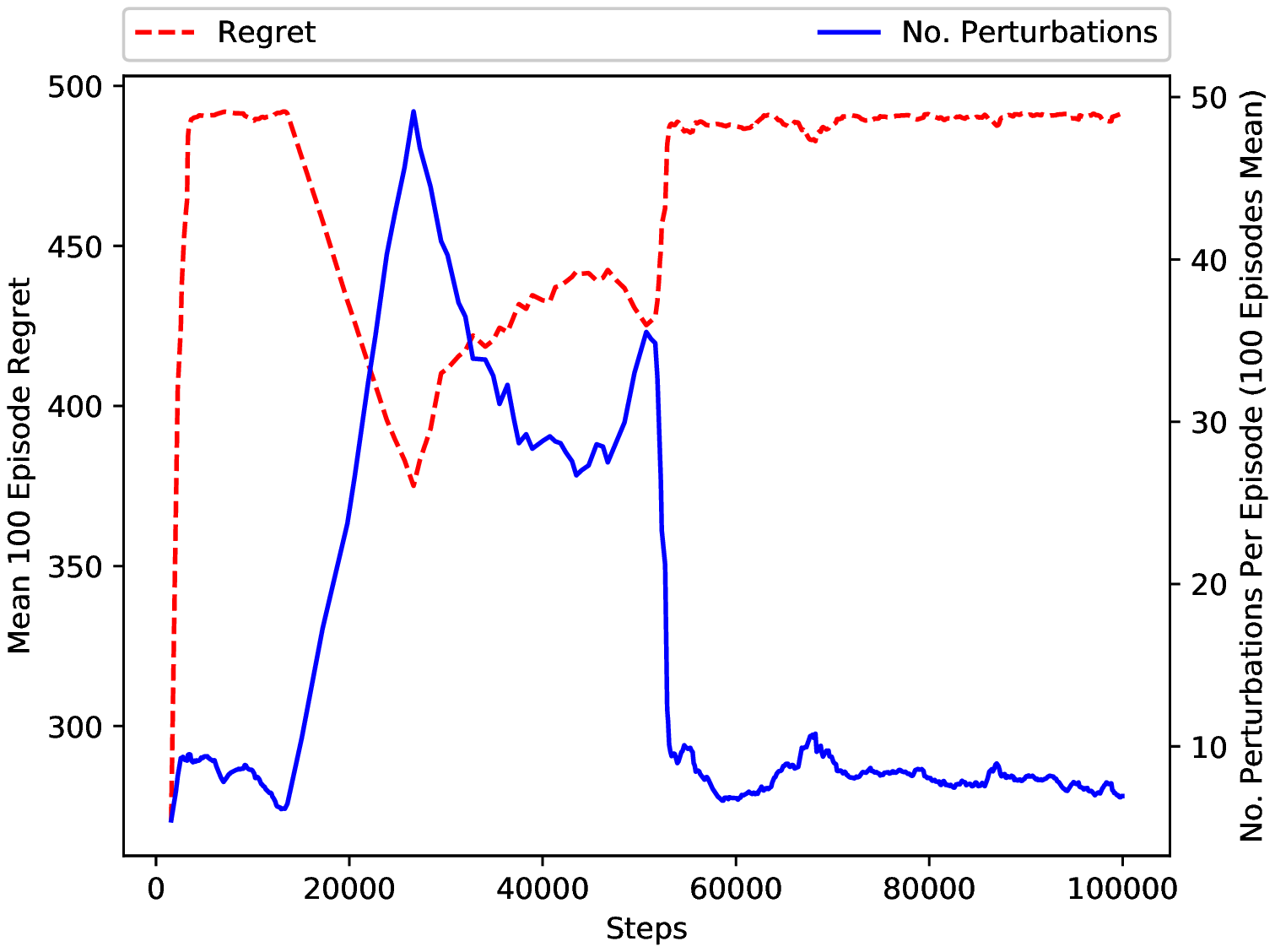}
	
	\caption{Adversarial Training Progress on A2C Policy with 1k demonstrations}
	
	\label{fig:A2C1k}
	
\end{figure}

\begin{figure}[H]
	
	\centering
	
	\includegraphics[width = 0.8\columnwidth]{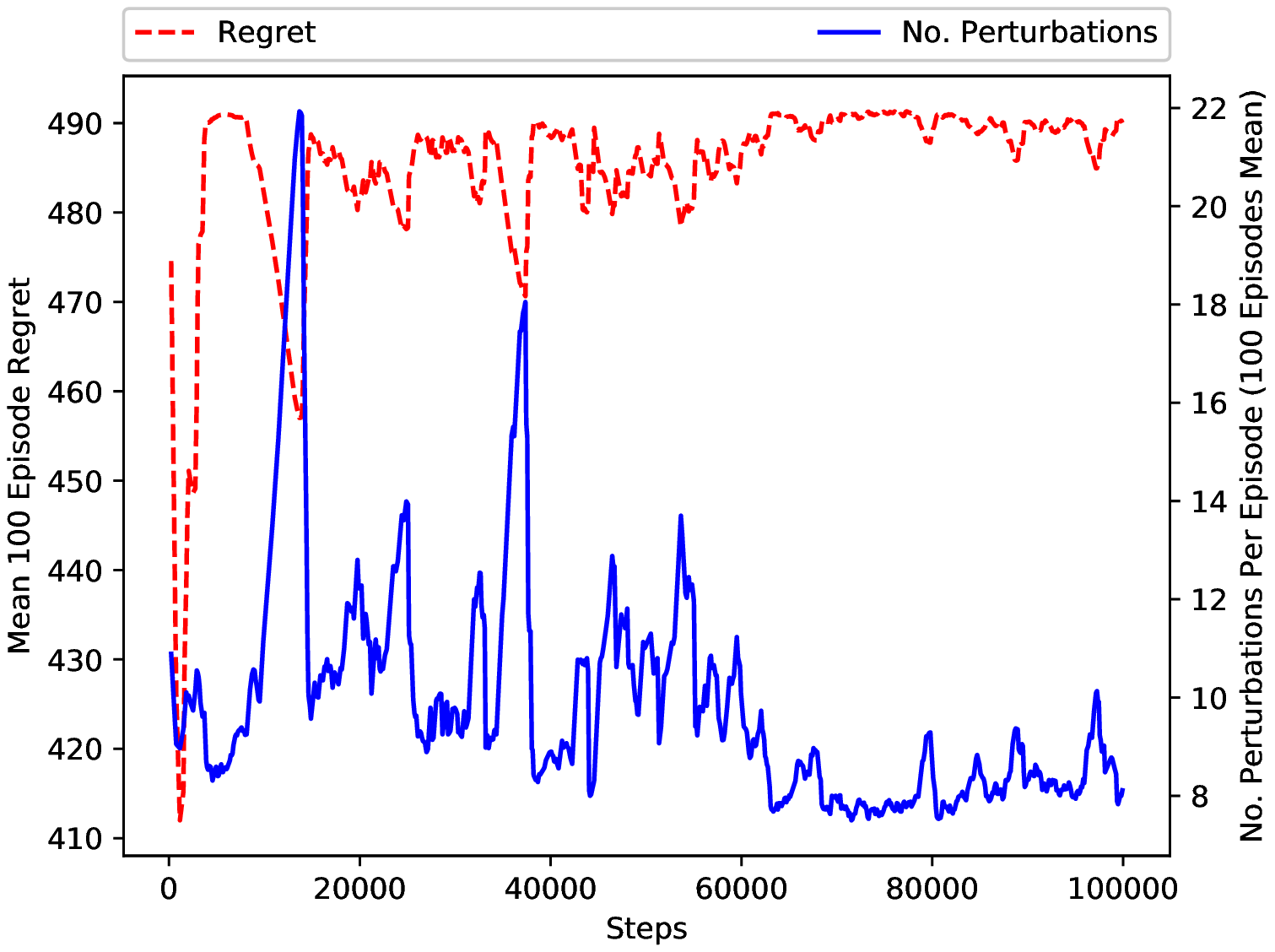}
	
	\caption{Adversarial Training Progress on PPO2 Policy with 5k demonstrations}
	
	\label{fig:PPO5k}
	
\end{figure}

\begin{figure}[H]
	
	\centering
	
	\includegraphics[width = 0.8\columnwidth]{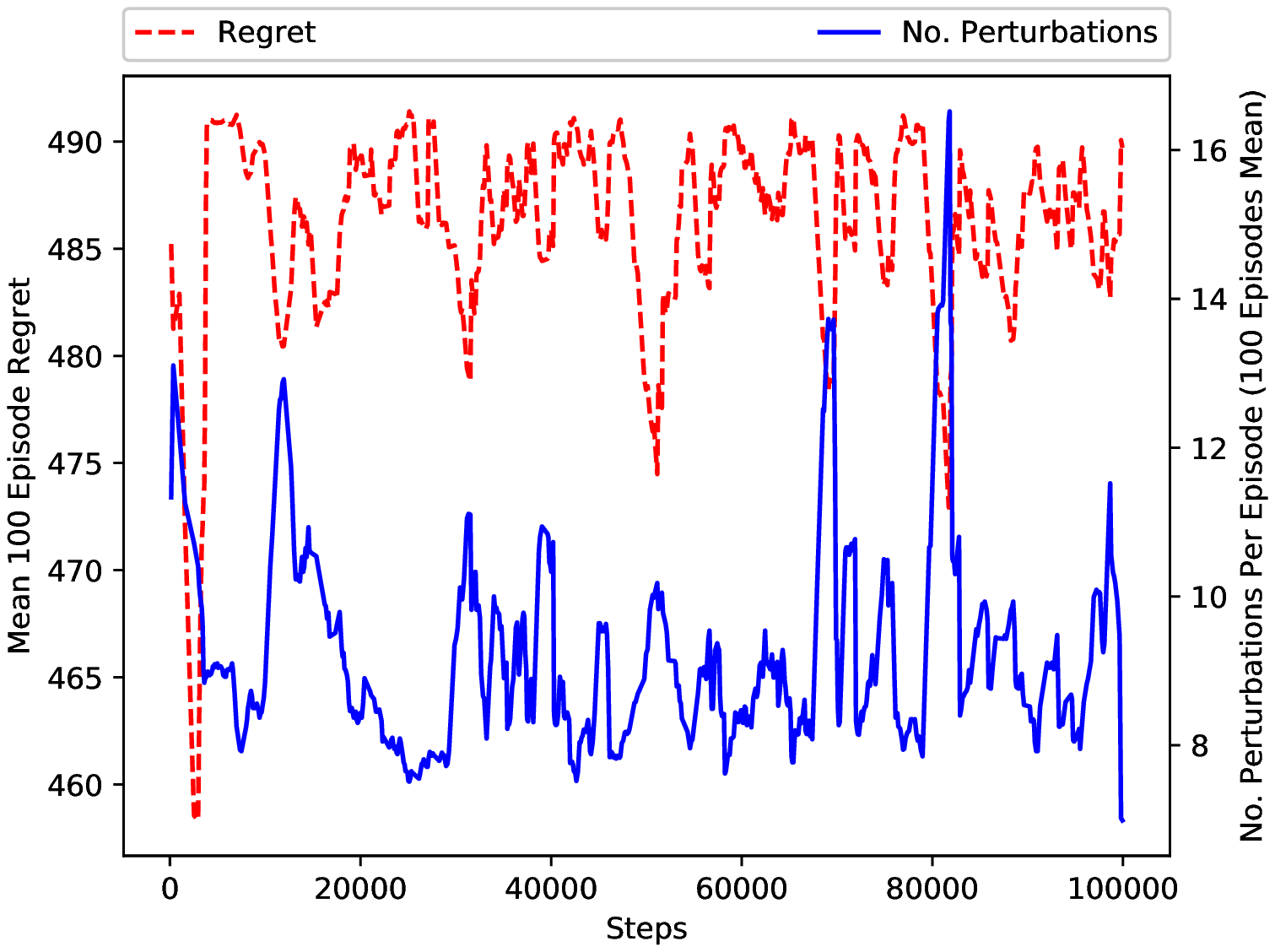}
	
	\caption{Adversarial Training Progress on PPO2 Policy with 2.5k demonstrations}
	
	\label{fig:PPO2k}
	
\end{figure}

With the imitated policies at hand, the next step is to train an adversarial policy for efficient perturbation of these targets. Figures \ref{fig:DQN5k} -- \ref{fig:PPO2k} present the results obtained from adopting the procedure presented in Algorithm \ref{alg:resilience} for this purposes. These results demonstrate that not only the limited training period is sufficient for obtaining an efficient adversarial policy, but also that launching efficient attacks remain feasible with relatively few observations (i.e., 2500 and 1000). However, the comparison of test-time performance of these policies (presented in table \ref{Table:TestTimeDQfD}) indicates that the efficiency of attacks decreases with lower numbers of observations.


\textbf{DQN:}

\textbf{A2C:}

\textbf{PPO2}

\begin{figure}[H]
	
	\centering
	
	\includegraphics[width = 0.8\columnwidth]{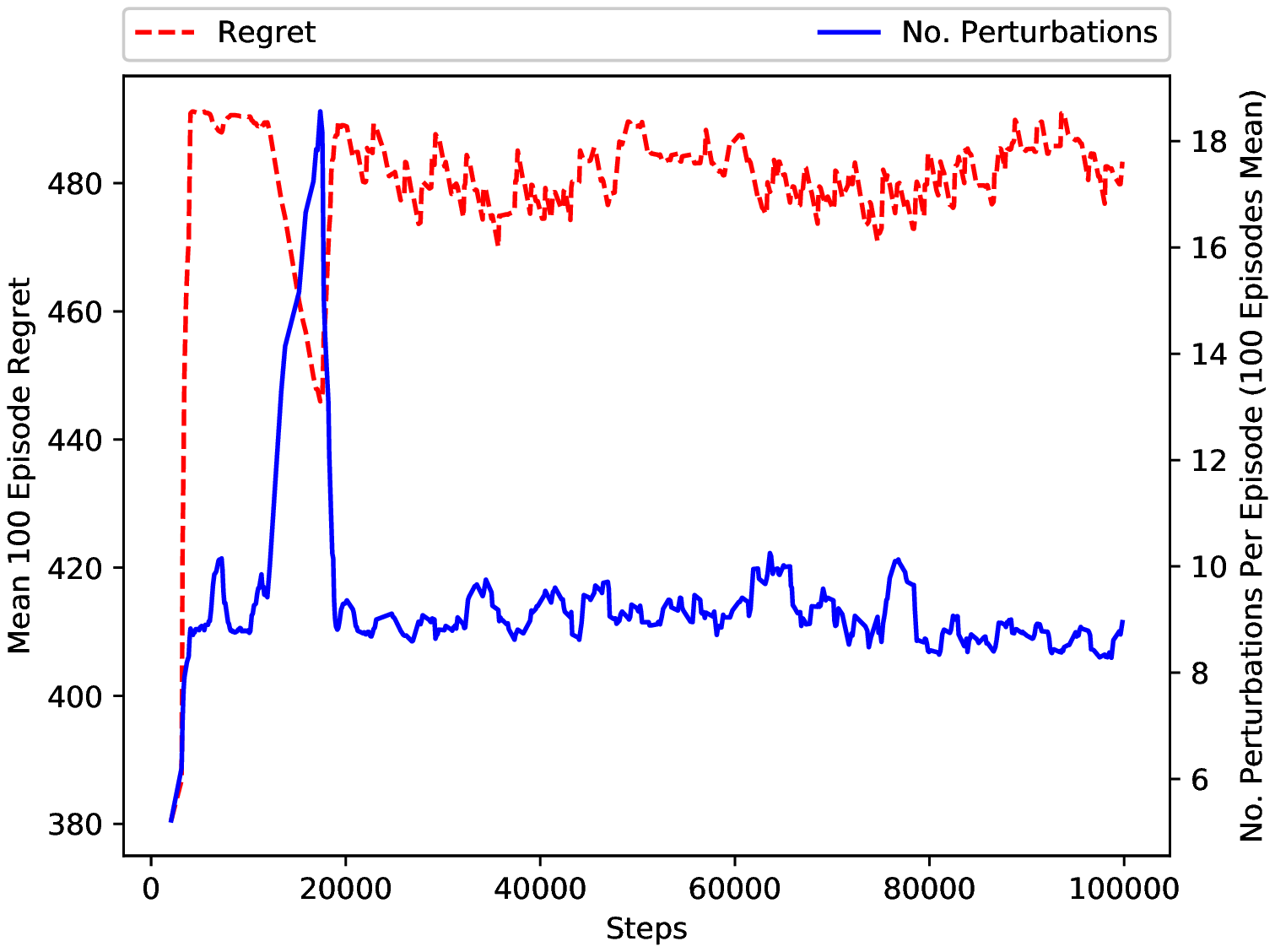}
	
	\caption{Adversarial Training Progress on PPO2 Policy with 1k demonstrations}
	
	\label{fig:PPO1k}
	
\end{figure}

\begin{table}[]
\centering
\begin{tabular}{|c|c|c|}
\hline
\textbf{Target Policy} & \textbf{Avg. Regret} & \textbf{Avg. No. Perturbations} \\ \hline
DQN-5k                 & 490.73               & 7.12                            \\ \hline
DQN-2.5k               & 488.12               & 8.09                            \\ \hline
DQN-1k                 & 486.37               & 10.55                           \\ \hline
A2C-5k                 & 490.88               & 8.48                            \\ \hline
A2C-2.5k               & 487.64               & 8.73                            \\ \hline
A2C-1k                 & 487.21               & 6.23                            \\ \hline
PPO2-5k                & 490.23               & 8.73                            \\ \hline
PPO2-2.5k              & 487.23               & 7.76                            \\ \hline
PPO2-1k                & 477.61               & 7.31                            \\ \hline
\end{tabular}
\caption{Comparison of Test-Time Performances of Adversarial Policies}
\label{Table:TestTimeDQfD}
\end{table}

\section{Transferability of Adversarial Example Attacks on Imitated Policies}
\label{sec:transfer}
It is well-established that adversarial examples crafted for a supervised model can be used to attack another model trained on a similar dataset as that of the original model\cite{liu2016delving}. Furthermore, Behzadan et al.\cite{behzadan2017vulnerability} demonstrate that adversarial examples crafted for one DRL policy can transfer to another policy trained in the same environment. Inspired by these findings, we hypothesize that adversarial examples generated for an imitated policy can also transfer to the original policy. To evaluate this claim, we propose the following procedure for black-box adversarial example attacks on DRL policies based on DQfD-based policy imitation: \begin{enumerate}
    \item Learn an imitation of the target policy $\pi$, denoted as $\Tilde{\pi}$.
    \item Craft adversarial examples for $\Tilde{\pi}$.
    \item Apply the same adversarial examples to the target's $\pi(s)$.
\end{enumerate}

\subsection{Experiment Setup}
We consider a set of targets consisting of the 9 imitated policies obtained in the previous section (i.e., DQN, A2C, PPO2, trained on each case of beginning with 5k, 2.5k, and 1k expert demonstrations). In test-time runs of each policy, we construct adversarial examples of each state against the imitated policy, using FGSM with perturbation step size $eps = 0.01$ and perturbation boundaries $[-5.0, 5.0]$. If such a perturbation is found, we then present it to the original policy. If the action selected by the original policy changes as a result of the perturbed input, then the adversarial example is successfully transferred from the imitated policy to the original policy.

\subsection{Results}
\begin{table}[]
\centering
\begin{tabular}{|c|c|}
\hline
\textbf{Target Policy} & \textbf{Avg. No. Successful Transfers Per Episode} \\ \hline
DQN-5k                 & 175.11                                             \\ \hline
DQN-2.5k               & 78.19                                              \\ \hline
DQN-1k                 & 3.30                                               \\ \hline
A2C-5k                 & 156.44                                             \\ \hline
A2C-2.5k               & 151.47                                             \\ \hline
A2C-1k                 & 21.58                                              \\ \hline
PPO2-5k                & 173.94                                             \\ \hline
PPO2-2.5k              & 112.96                                             \\ \hline
PPO2-1k                & 74.71                                              \\ \hline
\end{tabular}
\caption{No. of Successful Transfers Per Episode of Length 500 (100 Episode Mean)}
\label{Table:Transfer}
\end{table}
Table \ref{Table:Transfer} presents the number of successful transfers averaged over 100 consecutive episodes. These results verify the hypothesis that adversarial examples can transfer from an imitated policy to the original, thereby enabling a new approach to the adversarial problem of black-box attacks. Furthermore, the results indicate that the transferability improves with more demonstrations. This observation is in agreement with the general explanation of transferability: higher numbers of expert demonstrations decrease the gap between the distribution of training data used by the original policy and that of the imitated policy. Hence, the likelihood of transferability increases with more demonstrations. 


\section{Discussion on Potential Defenses}
\label{sec:discussion}
Mitigation of adversarial policy imitation is achieved by increasing the cost of such attacks to the adversary. A promising venue of research in this area is that of policy randomization. However, such randomization may lead to unacceptable degradation of the agent's performance. To address this issue, we envision a class of solutions based on the Constrained Randomization of Policy (CRoP). Such techniques will intrinsically account for the trade-off between the mitigation of policy imitation and the inevitable loss of returns. The corresponding research challenge in developing CRoP techniques is to find efficient and feasible constraints, which restrict the set of possible random actions at each state $s$ to those whose selection is guaranteed (or are likely within defined certainty) to incur a total regret that is less than a maximum tolerable amount $\Omega_{max}$. One potential choice of constraint is those applied to the $Q$-values of actions, leading to the technique detailed in Algorithm \ref{Algo:CRoP}. However, analyzing the feasibility of this approach will require the development of models that explain and predict the quantitative relationship between number of observations and accuracy of estimation. With this model at hand, the next step is to determine the saddle-point (or region) in the minimax settings of keeping the threshold $\Omega_{max}$ low, while providing maximum protection against adversarial imitation learning. This extensive line of research is beyond the scope of this dissertation, and is only introduced as a potential venue of future work to interested readers.  

\begin{algorithm}[H] 
\caption{Solution Concept for Constrained Randomization of Policy (CRoP)} 
\label{Algo:CRoP} 
\begin{algorithmic} 
    \REQUIRE state-action value function $Q(.,.)$, maximum tolerable loss $\Omega_{max}$, set of actions $A$
    \WHILE{Running}
        \STATE $s = env(t=0)$
        \FOR{each step of the episode}
            \STATE FeasibleActions $=\{\}$
            \STATE $a = \argmax_{a} Q(s,a)$
            \STATE Append $a$ to FeasibleActions
            \FOR{$a'\in A$}
                \IF{$Q(s,a) - Q(s,a') \geq \Omega_{max}$}
                    \STATE Append $a'$ to FeasibleActions
                \ENDIF
            \ENDFOR
            \IF {$\vert FeasibleActions\vert > 1$}
                \STATE $a \leftarrow random(FeasibleActions)$
            \ENDIF
            \STATE $s' = env(s, a)$
            \STATE $s \leftarrow s'$
        \ENDFOR
    \ENDWHILE
\end{algorithmic}
\end{algorithm}
\bibliographystyle{named}
\bibliography{ijcai19}

\begin{thebibliography}{}

\bibitem[\protect\citeauthoryear{Behzadan and
  Munir}{2017}]{behzadan2017vulnerability}
Vahid Behzadan and Arslan Munir.
\newblock Vulnerability of deep reinforcement learning to policy induction
  attacks.
\newblock In {\em International Conference on Machine Learning and Data Mining
  in Pattern Recognition}, pages 262--275. Springer, 2017.

\bibitem[\protect\citeauthoryear{Hester \bgroup \em et al.\egroup
  }{2018}]{hester2018deep}
Todd Hester, Matej Vecerik, Olivier Pietquin, Marc Lanctot, Tom Schaul, Bilal
  Piot, Dan Horgan, John Quan, Andrew Sendonaris, Ian Osband, et~al.
\newblock Deep q-learning from demonstrations.
\newblock In {\em Thirty-Second AAAI Conference on Artificial Intelligence},
  2018.

\bibitem[\protect\citeauthoryear{Hussein \bgroup \em et al.\egroup
  }{2017}]{hussein2017imitation}
Ahmed Hussein, Mohamed~Medhat Gaber, Eyad Elyan, and Chrisina Jayne.
\newblock Imitation learning: A survey of learning methods.
\newblock {\em ACM Computing Surveys (CSUR)}, 50(2):21, 2017.

\bibitem[\protect\citeauthoryear{Liu \bgroup \em et al.\egroup
  }{2016}]{liu2016delving}
Yanpei Liu, Xinyun Chen, Chang Liu, and Dawn Song.
\newblock Delving into transferable adversarial examples and black-box attacks.
\newblock {\em arXiv preprint arXiv:1611.02770}, 2016.

\bibitem[\protect\citeauthoryear{Piot \bgroup \em et al.\egroup
  }{2014}]{piot2014boosted}
Bilal Piot, Matthieu Geist, and Olivier Pietquin.
\newblock Boosted bellman residual minimization handling expert demonstrations.
\newblock In {\em Joint European Conference on Machine Learning and Knowledge
  Discovery in Databases}, pages 549--564. Springer, 2014.

\bibitem[\protect\citeauthoryear{Tram{\`e}r \bgroup \em et al.\egroup
  }{2016}]{tramer2016stealing}
Florian Tram{\`e}r, Fan Zhang, Ari Juels, Michael~K Reiter, and Thomas
  Ristenpart.
\newblock Stealing machine learning models via prediction apis.
\newblock In {\em USENIX Security Symposium}, pages 601--618, 2016.

\end{thebibliography}
\end{document}